\documentclass{article}
\usepackage{arxiv}
\usepackage[utf8]{inputenc} 
\usepackage[T1]{fontenc}    
\usepackage{hyperref}       
\usepackage{url}            
\usepackage{booktabs}       
\usepackage{amsfonts}       
\usepackage{nicefrac}       
\usepackage{microtype}      
\usepackage{lipsum}
\usepackage{graphicx}
\graphicspath{ {./images/} }

\usepackage{times}
\usepackage{latexsym}
\usepackage[T1]{fontenc}
\usepackage[utf8]{inputenc}
\usepackage{microtype}
\usepackage{inconsolata}
\usepackage{longtable}
\usepackage{adjustbox}
\usepackage{hyperref}
\usepackage{amssymb,amsthm,amsmath}
\usepackage{xcolor,paralist,hyperref,titlesec,fancyhdr,etoolbox}

\title{Multi-head Sequence Tagging Model for Grammatical Error Correction}

\author{Kamal Al-Sabahi \\
University of Technology and Applied Sciences\\
Sultanate of Oman\\
\And
Kang Yang \\
National University of Defense Technology \\
China\\
  \And
Wangwang Liu \\
CVTE Research \\
China \\
  \And
Guanyu Jiang \\
Dalian University of Technology \\
China \\
  \And
Xian Li \\
CVTE Research \\
China \\
  \And
Ming Yang \\
CVTE Research \\
China \\}

\begin{document}
\maketitle
\begin{abstract}To solve the Grammatical Error Correction (GEC) problem , a mapping between a source sequence and a target one is needed, where the two differ only on few spans.  For this reason, the attention has been shifted to the non-autoregressive or sequence tagging models. In which, the GEC has been simplified from Seq2Seq to labeling the input tokens with edit commands chosen from a large edit space. Due to this large number of classes and the limitation of the available datasets, the current sequence tagging approaches still have some issues handling a broad range of grammatical errors just by being laser-focused on one single task. To this end, we simplified the GEC further by dividing it into seven related subtasks: Insertion, Deletion, Merge, Substitution, Transformation, Detection, and Correction, with Correction being our primary focus. A distinct classification head is dedicated to each of these subtasks. the novel multi-head and multi-task learning model is proposed to effectively utilize training data and harness the information from related task training signals. To mitigate the limited number of available training samples, a new denoising autoencoder is used to generate a new synthetic dataset to be used for pretraining. Additionally, a new character-level transformation is proposed to enhance the sequence-to-edit function and improve the model's vocabulary coverage. Our single/ensemble model achieves an F0.5 of 74.4/77.0, and 68.6/69.1 on BEA-19 (test) and CoNLL-14 (test) respectively. Moreover, evaluated on JFLEG test set, the GLEU scores are 61.6 and 61.7 for the single and ensemble models, respectively. It mostly outperforms recently published state-of-the-art results by a considerable margin.
\end{abstract}
\section{Introduction}
\label{sec:introduction}
Grammatical Error Correction (GEC) is the task of detecting and correcting different kinds of errors, such as punctuation, spelling, grammatical, and word choice \cite{Jia2013}. In recent years, it has been the subject of many modeling efforts due to its ability to improve the grammaticality and readability of user-generated texts \cite{Rothe2021}. Most of the previously proposed approaches for GEC view the task as monolingual text-to-text rewriting or machine translation \cite{Bryant2019}, \cite{Kiyono2019}, \cite{Zhao2019}, \cite{Rothe2021}, \cite{Stahlberg2021} and \cite{Lichtarge2019}. While these models can handle the full dependency between outputs \cite{al2018bidirectional}, some serious issues still need to be addressed before we get a GEC model that matches the human performance. GEC is different from machine translation in the sense that it only changes several words of the source sentence. Previous works \cite{Awasthi2019}, \cite{Omelianchuk2020} have shown that during inference, the main efficiency bottleneck of Seq2Seq models returns to the fact that most of the decoding steps are spent copying grammatically correct tokens from the original sentences to target ones \cite{Li2022} \cite{Kamal2018, mohsen2020hierarchical}. The efficiency of GEC models can be improved by saving that time. Moreover, Seq2Seq methods typically require large training sets to work well, which are scarce and challenging. 

Nowadays, sequence labeling approaches have been used to approach the GEC problem. It has been simplified from sequence generation to sequence labeling, predicting edits instead of tokens. This allows parallel processing at the inference time instead of sequential decoding. However, the current sequence labeling/tagging-based approaches still have some issues that need to be addressed. One of which is the complication of handling a large number of edit classes under the limited availability of training data. In this work, we tried to overcome this issue by leveraging a multi-task learning approach. 

Multi-task learning (MTL) aims to learn multi-tasks simultaneously with a shared model \cite{Changpinyo2018},\cite{10181308}.  It is originally inspired by human learning. In which, we intend to apply prior knowledge learned by related tasks to master more complex ones  \cite{Schröder2020, Wang2021}. In the context of machine learning, MTL has the advantage of improving data efficiency, using the shared representation to reduce overfitting, and leveraging auxiliary information for fast learning \cite{Crawshaw2020}.

\begin{figure*}[t]
	\centering
	\includegraphics[width=\textwidth,height=\textheight,keepaspectratio]{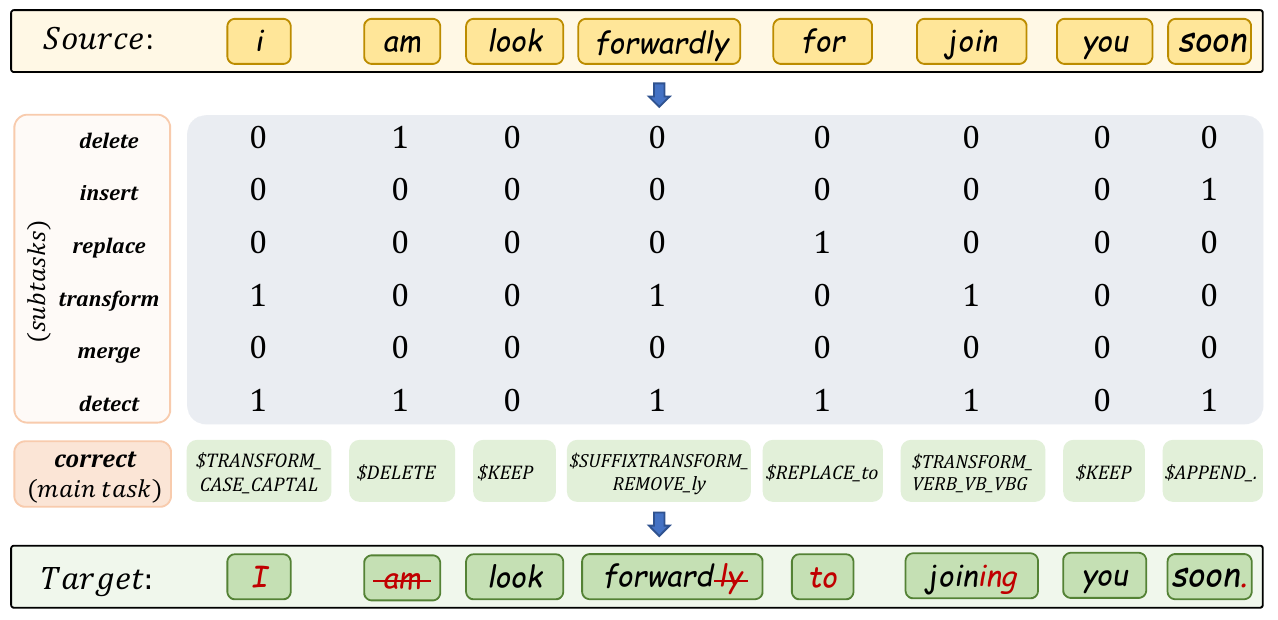}
	\caption{Labels for each GEC subtask generated from the same source sentence, an example.}
	\label{tab:heads}
\end{figure*}

The shared representation between two or more related tasks can improve the generalization of the original task \cite{Wang2021}. The inductive bias provided by the auxiliary tasks leads to solutions that generalize better since the model prefers hypotheses that explain more than one task \cite{Ruder2017}, \cite{Zhang2021}.  A recent work \cite{Fynn2020} found auxiliary tasks with compact and uniform label distributions to be preferable for sequence tagging problems. The auxiliary data can be incorporated into the training to boost the main task performance. In this work, we propose novel multi-head model to improve the efficiency and the performance of the GEC model by dividing the task into seven subtasks: Insertion, Deletion, Merge, Substitution, Transformation, Detection and Correction, shown in Figure \ref{tab:heads}. Since previous works like \cite{Awasthi2019}, \cite{Lichtarge2019}, \cite{Lichtarge2020} \cite{Omelianchuk2020}, and \cite{Rothe2021} have proven the efficiency of multistage training, our GEC sequence tagging pipeline consists of three training stages: pretraining on synthetic data, fine-tuning on an errorful parallel corpus, and finally, fine-tuning on a combination of errorful and error-free parallel corpora.

The contributions of this work are three-fold. 
\begin{enumerate}
	\item A novel GEC neural architecture is proposed in which the GEC tagging problem is simplified into several subtasks learned jointly. 
	\item To mitigate the lack of labeled data, we use a denoising auto-encoder model to convert huge unlabeled datasets into artificially labeled ones. 
	\item We propose a character-level transformation to mitigate the effect of the limited vocabulary size. 
 \item We released the code \footnote{\url{https://github.com/alsabahi2030/MH-GEC}}, the generated pretraining dataset \footnote{\url{https://drive.google.com/file/d/1ZYxWra8h_nUqWydmRIzgjUiJlo_2eTgo/view?usp=drive_link}} and the code \footnote{\url{https://github.com/alsabahi2030/Synthetic-Data-Generation}} of generating it.
\end{enumerate}
\section{Related Work}
\label{sec:related}
\textbf{Sequence Tagging Approaches} To overcome the previously mentioned issues in Seq2Seq models, attention has been shifted to applying sequence labeling-based approaches to the GEC task \cite{Awasthi2019} \cite{Li2022} , \cite{Omelianchuk2020}, \cite{Rothe2021}, \cite{Stahlberg2021}. Some of which are language-dependent approaches, like PIE \cite{Awasthi2019} and GECToR \cite{Omelianchuk2020}. A sequence of token-level edits is predicted out of a number of manually designed edit operations like changing the form of a verb (e.g., changing a 3rd person singular present Verb (VBZ) to the past tense form (VBD)) or replacing a preposition (e.g., in/on), and etc. One of the shortcomings of those models is that they see a large variety of span annotations and need to learn them during training. The large number of classes makes it difficult for the model to learn them without any other assistance from other related tasks.  In our work, we leverage the multi-task learning setup to boost the performance. Moreover, the model’s robustness would be improved even if the detected spans during inference are not exactly accurate; the correction model, the main task, will not easily fail. 

Another recent work by \cite{didenko_redpennet_2023} combines autoregressive and sequence tagging techniques to streamline semi-autoregressive Sequence-To-Edits models for text editing, yet this introduces complexity into the model architecture. In contrast, our multi-head sequence tagging model deconstructs GEC into simpler, distinct subtasks, offering a less complex and more focused solution. One more study introduced by \cite{sorokin_improved_2022}, in which a two-stage reranking method is used, initially generating edits and then classifying them as correct or incorrect. Unlike their model which relies on reranking for refinement, our approach directly addresses a wide array of grammatical errors through a multi-head sequence tagging model, eliminating the need for a reranking phase and maintaining a streamlined architecture.
 
\textbf{Multi-Task Approaches} In MTL, a shared representation is jointly learned from multiple tasks \cite{Fynn2020}, \cite{Wang2021}. Theoretical results have shown that the joint training scheme with MTL is more sample efficient than single-task learning, at least under certain assumptions of task relatedness, linear features, and model classes \cite{Wang2021}. A work proposed by \cite{Fynn2020} tried to study the influence of auxiliary tasks on multi-task learning for sequence tagging problems. They concluded that auxiliary datasets with specific degree of similarity with the main task has increased the latter performance. In the context of GEC, the work of \cite{Zhao2019} is the only work that mentions the multi-task learning for GEC. They added token-level and sentence-level multi-task learning for the GEC task. However, their model is a Seq2Seq model that inherited all the Seq2Seq  issues mentioned earlier in this section.

\textbf{Data Augmentation} The focus of a work by \cite{Rothe2021}, is to train a large universal model for multilingual GEC using a large-scale multilingual language model called mT5 \cite{Xue2021}. Another interesting work was proposed by \cite{Stahlberg2021}. In which, they tried to solve the different error distribution using error type tags from ERRANT \cite{Bryant2017} that is used to control the synthetic data generation. However, existing data augmentation can only improve the GEC model’s handling of specific types of errors since they are still far from generating errors with the same distribution as the human one  \cite{Wan2020}. In this work, we tried to generate human-like errors by applying several noising techniques that follow the error distribution of BEA-19 dev \cite{Bryant2019}.
\section{Multi-head Sequence Tagging model}
The GEC problem is formulated as labeling the input sequence $x_1$,…,$x_n$ with edit tags $e_1$,…,$e_n$. Converting the parallel dataset to this format needs a careful design of the Seq2Edit function. The function takes a pair of sequences (source $x$, target $y$) as input, and outputs the corresponding pair of source and edits (source  $x$,edits $e$). To increase the coverage of GEC on limited vocabulary size, we use two kinds of transformations, token-level, and character-level, that can be applied to the most common English grammatical errors. The target sentences can be recovered by applying those transformations to the corresponding source sentences. For all other auxiliary tasks, a binary sequence tagging approach is used, in which tag “0” is assigned when the token is correct, and the tag “1” is assigned when the token is incorrect and needs to be edited, as shown in Table \ref{tab:heads}. In the following subsections, we will discuss the Seq2Edit function and the two types of transformations, Section \ref{sec:seqedit} .

\subsection {The Seq2Edit Function}
\label{sec:seqedit}
The Seq2Edit function is integral to our GEC model, transforming the source sentence into a sequence of edit commands. This process involves two main steps:
\begin{enumerate}
\item[(a)] Token Mapping: Employing a methodology akin to that of \cite{Omelianchuk2020}, each token from the source sentence is mapped to a corresponding subsequence in the target sentence. This foundational step establishes a clear relationship between the erroneous input and the corrected output.
\item[(b)] Transformation Identification: For each mapped pair, the function identifies the necessary token-level or character-level transformations, as discussed in Section \ref{sec:tokenlevel} and Section \ref{sec:charlevel}. These are then translated into edit commands, guiding the model in its correction process.
\end{enumerate}

In a more technical sense, given an input sequence $x$=($x_1$,…,$x_n$) and an output sequence $y$=($y_1$,…,$y_m$), $(x;y)\in D$, where the input sequence's length $m$ may not be equal to  the output sequence's length $n$, our goal is to derive a sequence of edit operations $e$=($e_1$,…,$e_n$), $e \in \upsilon$. This process ensures that applying each edit operation $e_i$ on the corresponding source token $x_i$ recovers the output sequence $y$. The Seq2Edit takes $(x;y)$ pair from $D$ as input and outputs a sequence $e$ of edit operations from an edit space $\upsilon$ where $e$ is of the same length as $x$ in spite of $x$ and $y$ being of different lengths, equation \eqref{eq:seq2edit}. Hence, the insert edit is merged with its preceding edit operations creating a compound append or replace operation.
\begin{equation}        
	(x,e)=Seq2Edit( x,y)            
	\label{eq:seq2edit}
\end{equation}
where $x$=($x_1$,…,$x_n$) represents the input sequence and $y$=($y_1$,…,$y_m$) indicates the output sequence taken from a document $D$. Additionally, $e$=($e_1$,…,$e_n$), $e \in \upsilon$, represents the corrsponding sequence of edits from the edit space $\upsilon$.
\subsubsection{Token-level transformations}
\label{sec:tokenlevel}
Token-level transformations perform the most common token-level edit operations, such as: keep the current token unchanged (tag \texttt{$KEEP$}), delete current token (tag \texttt{$DELETE$}), append a new token  $t_i$ next to the current token $x_i$ (tag \texttt{$APPEND_{t_i}$}) or replace the current token $x_i$ with another token $t_j$ (tag \texttt{$REPLACE_{t_j}$}). Following the work of \cite{Omelianchuk2020}, we use a token-independent transformation that includes, \texttt{$KEEP$} and \texttt{$DELETE$}, and a token-dependent transformation which includes, \texttt{$REPLACE_{token}$} and  \texttt{$APPEND_{token}$}. We have added two edit types to this category. The first one merges the current token and the next token into a single one (\texttt{$MERGE$} tags) and the second splits the current token into two new tokens (\texttt{$SPLIT$} tags) as shown in Table \ref{tab:basic_edit_trans}, \ref{sec:appendix}.
\subsubsection{Character-level Transformations }
\label{sec:charlevel}
To improve the coverage of the vocabulary, we introduce a new character-level transformation, in which a source token is transformed to a target one by deleting, replacing, appending, merging, splitting, or changing the case of one or more characters of the source token. This enables a single transformation tag to represent a broad number of cases that were dealt separately in previous works like \cite{Omelianchuk2020}. For example, change the case of the current token (\texttt{$CASE$} tags). Moreover, some of those transformations include changing the grammatical properties of the tokens, such as converting singular nouns to plurals and vice versa or even changing the form of regular/irregular verbs to express a different number or tense. Transformations perform inflections like adding suffix \texttt{s}, \texttt{d}, \texttt{es}, \texttt{ing}, \texttt{ed}, changing suffix \texttt{s} to \texttt{ing}, or \texttt{d} to \texttt{s}, etc, as shown in Tables \ref{tab:edit_trans} and Tables \ref{tab:suffix_trans_append},\ref{tab:suffix_trans_remove}, \ref{tab:suffix_trans_replace}, \ref{sec:appendix}.
\subsection{Denoising auto-encoder}
\label{sec:DAE}
GEC problem suffers from a lack of clean parallel datasets \cite{Lichtarge2019}. The publicly available annotated datasets are small or noisy \cite{Lichtarge2020}. Over the years, several approaches have been proposed to tackle the GEC data scarcity mostly by generating synthetic training data. It has been shown that pretraining GEC on those artificially generated data has boosted the performance \cite{Rothe2021}. In this work, we leverage huge unlabeled datasets by corrupting them following a realistic noising scenario that can produce ungrammatical sentences given clean sentences with error distributions matching a given development set. The following datasets are used: One-Billion-Word dataset \cite{Chelba2013}, Gutenberg \footnote{\url{https://drive.google.com/uc?id=0B2Mzhc7popBga2RkcWZNcjlRTGM}}, Tatoeba \footnote{\url{http://downloads.tatoeba.org/exports/sentences.tar.bz2}}, WikiText-103 \footnote{\url{https://s3.amazonaws.com/research.metamind.io/wikitext/wikitext-103-v1.zip}}, and Question Answering \footnote{\url{https://rajpurkar.github.io/SQuAD-explorer/}}. We corrupt each sentence using a combination of the following operations:
\begin{itemize}
	\item \textbf{Token-based and Type-based methods:}
	This approach is an extension of the one used by \cite{KakaoBrain2019} in which two types of errors are generated, Token-based and type-based errors. In the former type, human edits are extracted from the annotated BEA train  dataset using ERRANT toolkit \cite{Bryant2017} to build a dictionary of token edits for common errors. Then apply the common edits in reverse. In the latter, they generated new errors based on token type including prepositions, adjective nouns, and verbs. In other words, a replacement from same type is done to generate the error. For example, a proposition is replaced with another preposition (e.g., of → off), a noun with its singular/plural form, a verb with one of its inflected forms, etc.
	\item \textbf{Injecting sophisticated errors:}
	Sophisticated errors are generated by switching, inserting, deleting, or replacing common bi-grams/tri-grams (e.g., I am → am I, United State → State United). In addition, more errors are obtained by switching, replacing, or deleting common character sequences (e.g., ea → ae, ion → oin) and changing between error types, such as replacing an adjective with an adverb (e.g., slow → slowly, fast → fastly), as shown in Table \ref{tab:error_pattterns}, \ref{sec:appendix}.
\end{itemize}
\subsection{Model}
\label{sec:baseArch}
Our base model is just our implementation of \cite{Omelianchuk2020}. The model consists of a deep bidirectional transformer \cite{Vaswani2017} as an encoder, providing a contextual encoding for each $x_i$, and a classifier for GEC sequence tagging. The encoder is made up of a pre-trained BERT-like Transformer, such as XLNet, RoBERTa, BERT, etc. We always use cased pretrained transformers in their base configurations.  On the top of the encoder, a linear layer is used as a classification head with a \texttt{SoftMax} layer on the top. The classifier will determine whether the sentence is correct. If not, it will predict the specific edit to correct the grammatical error, equations \eqref{eq:encoder}, \eqref{eq:softmax}, and \eqref{eq:ehat}. 
\begin{equation}
	H = [h_1,h_2, ... ,h_{n+1}] = encoder(x)                              
	\label{eq:encoder}
\end{equation}
\begin{equation}
	P(e_i\mid x)=softmax(W_e h_i) , \  e \in \upsilon
	\label{eq:softmax}
\end{equation}
\begin{equation}
	\hat{e} = argmax_e \ p(e \mid x,\theta)  
	\label{eq:ehat}
\end{equation}
\begin{equation}
	\hat{y} = Edit2Seq(x,\hat{e}), \ \hat{e}\in \upsilon
	\label{eq:yhat}
\end{equation}
\noindent
Where $H$ is the output of the encoder. Each $h_i$ is a vector that represents the $i^{th}$ token in the input sequence and $W_e$ and $\theta$ are learnable parameters. Additionaly, $Edit2Seq()$ is the function that applies  the edit $\hat{e}$ to the input $x$ to get the target sequence $\hat{y}$ and  $\upsilon$ is the edit space.
\subsubsection{Multi-Task Learning}
We propose a multi-head classification model with $N$ heads trained over $N$ training subtasks. For each token in the source sentence, a label is assigned indicating whether this token is incorrect, should be deleted, append to it, replaced, merged with the next one, or transformed to another form by adding/removing suffixes. Assuming that each source token $x_i$ can be aligned with a target token $y_j$, we define that the source token is correct if $x_i= y_j$, and wrong otherwise.

\begin{figure*}[t]
	\centering
	\includegraphics[width=\textwidth,height=\textheight,keepaspectratio]{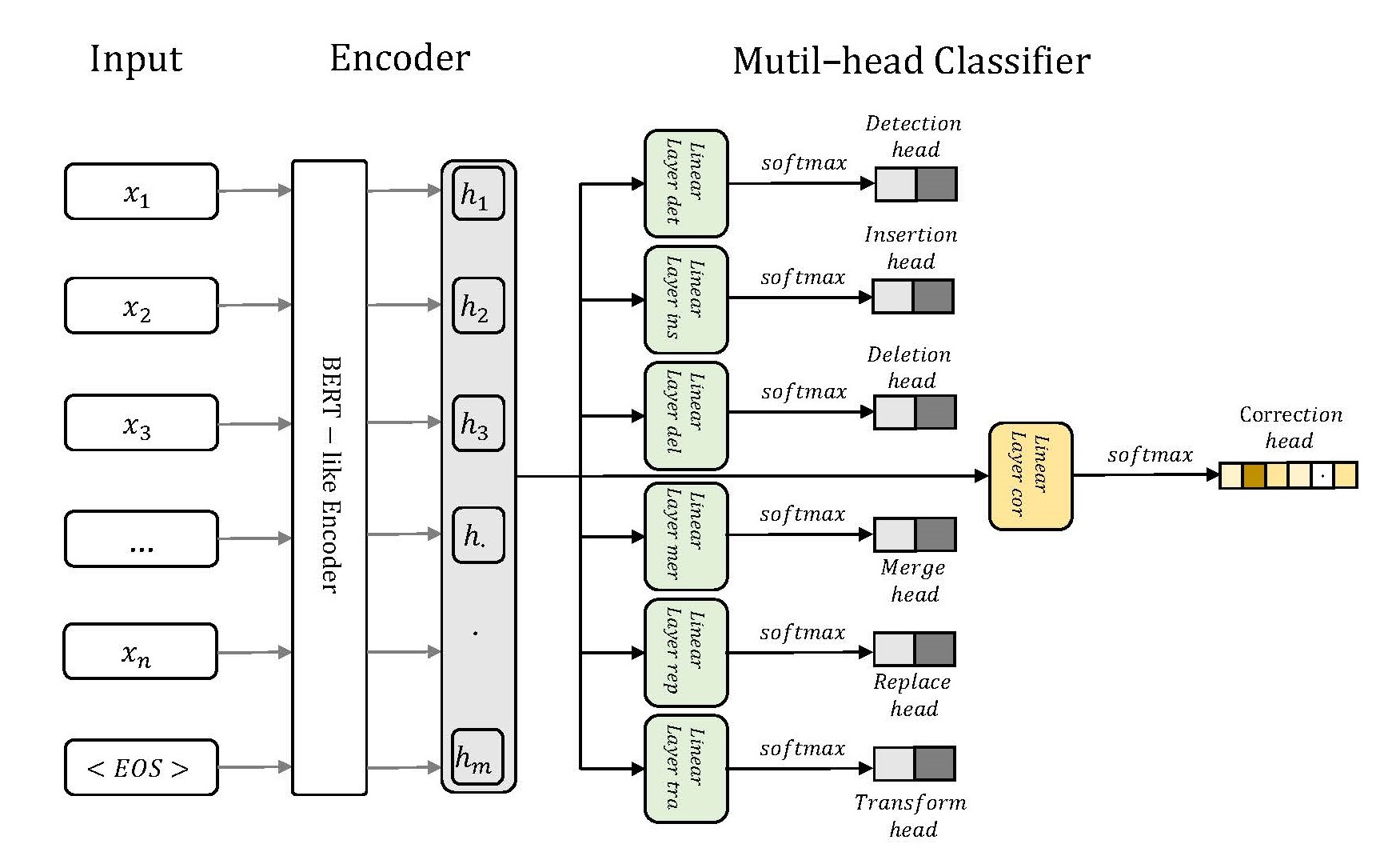}
	\caption{The proposed multi-head classifier (Multi-head7). The left part contains the shared pretrained layers (BERT-like encoder). And the right part contains linear layers that are task-specific and can be learned separately for the corresponding task. On the top of each linear layer, there is a \texttt{SoftMax} layer for classification.}
	\label{fig:model_figure}
\end{figure*}

Each token’s label is predicted by passing the final state of the  BERT-like encoder through a \texttt{SoftMax} after an affine transformation, as shown in equations \eqref{eq:delete}, \eqref{eq:insertion}, \eqref{eq:substitute}, \eqref{eq:merge}, \eqref{eq:trans}, \eqref{eq:detect},\eqref{eq:correct}. Figure \ref{fig:model_figure} depicts the basic components of our proposed framework.

\begin{enumerate}
	\item \textbf{Deletion classifier:} 
	The model scans over the input tokens (except for the boundaries) and predicts “delete” (1) or “keep” (0) for each token position.
	\begin{equation}
		P(D_i \mid x)=softmax(W_d h_i ) 
		\label{eq:delete}
	\end{equation}
	\item \textbf{Insertion Classifier:}
	The model predicts whether there is a missing token to be added at that position by assigning “append” (1) or “keep” (0) for each token position.
	\begin{equation}
		P(A_i\mid x)=softmax(W_a h_i )
		\label{eq:insertion}
	\end{equation}
	\item	\textbf{Substitution Classifier:}
	This classifier predicts whether there is a substitution “replace” (1) or there is no “keep” (0) at the current position.
	\begin{equation}
		P(R_i\mid x)=softmax(W_r h_i )
		\label{eq:substitute}
	\end{equation}
	\item	\textbf{Merge classifier:}
	This classifier predicts whether the current and the next token should be merged into one, "merge" (1) or "keep" (0).		
        \begin{equation}
		P(M_i\mid x)=softmax(W_m h_i )
		\label{eq:merge}
	\end{equation}
	\item	\textbf{Transformation classifier:}
	This classifier predicts whether a transformation is needed at the current position to convert the source token to the target one. 	
	\begin{equation}
		P(T_i\mid x)=softmax(W_t h_i )
		\label{eq:trans}
	\end{equation}
	\item	\textbf{Detection classifier:}
	This classifier predicts whether the current token is correct or incorrect. This information can be used for inference tweaking where we ignore the edits when the error probability is not high enough.	
	\begin{equation}
		P(B_i\mid x)=softmax(W_b h_i )
		\label{eq:detect}
	\end{equation}
	\item	\textbf{Correction Classifier:}
	This is the classifier of the main task.  It predicts the token-level and character-level transformations to recover the target text by applying them to the source tokens. 
	\begin{equation}
		P(C_i\mid x)=softmax(W_c h_i )
		\label{eq:correct}
	\end{equation}
\end{enumerate}
\noindent
where $W_d$, $W_a$, $W_r$, $W_m$, $W_t$, $W_b$ $\in$ $R^{2 \times d_{model}}$ 
denote the learnable parameters for delete, insert, replace, merge, transform, and detect heads, respectively, $i$=1,…$n$, and $W_c \in R^{\lvert \upsilon \rvert \times d_{model}}$ denotes the parameters for the correction head.

The multi-head model is trained to minimize the following loss, equation \eqref{eq:loss}:
\begin{multline}
	L_{total}=l_{c}+ \lambda l_{d}+\lambda l_{a}+\lambda l_{r}+	\lambda l_{t}+\lambda l_{m} + \lambda l_{b}
	\label{eq:loss}
\end{multline}
where $\lambda\in[0,1]$ and $l_{c}$, $l_{d}$, $l_{a}$, $l_{r}$, $l_{t}$, $l_{m}$, and $l_{b}$ are the individual losses of the classification heads: correct, delete, insert, replace, transform, merge, and detect respectively. 
\subsubsection{Iterative Refinement}
We follow the iterative correction approach of \cite{Awasthi2019}. The inference capacity of the parallel model is increased by iteratively inputting the model’s own output for further refinement. We iteratively refine predictions to capture dependencies using the GEC sequence tagger to tag the now modified sequence and apply the corresponding transformations on the new tags, which changes the sentence further.
\section{Experiments}
\subsection{Datasets}
\label{sec:data}
In this subsection, we describe the datasets used in our experiments. Our study leverages the One-Billion-Word dataset, Gutenberg, Tatoeba, WikiText-103, and Question Answering datasets for pretraining, as described in Section \ref{sec:DAE}. The public version of the Lang-8 corpus  \cite{Mizumoto2011}, NUCLE \cite{Dahlmeier2013}, the FCE corpus \cite{Yannakoudakis2011} and the Cambridge English Write \& Improve training split \cite{Bryant2019} described in the BEA-19 shared task (BEA-19 train) \cite{Bryant2019} are combined and used for the first finetuning stage. For the second finetuning stage, we used only W\&I+LOCNESS \cite{Bryant2019}, for domain adaptation. These datasets provide a comprehensive range of grammatical constructions and error types, ensuring the robustness and generalizability of our model.
The proposed models are evaluated on standard benchmarks from CoNLL-14 \cite{Ng2014}, BEA-19 test \cite{Bryant2019}, and JFLEG test set \cite{napoles-etal-2017-jfleg}  with their official evaluation scripts (m2scorer2 for CoNLL-14, ERRANT \cite{Bryant2017} for BEA-19 and GLEU score for JFLEG). The CoNLL-13 \cite{Ng2013} is used as a validation set during pretraining and BEA-19 dev \cite{Bryant2019} is used as a validation set for the two stages of finetuning.

\subsection{Vocabulary}
At its core, our model operates as a classifier, designed with a comprehensive list of classes that correspond to various edit operations taken for an extensive edit space $\upsilon$. The edit space $\upsilon$ that corresponds to our default tagset (vocabulary size = 5000) consists of 1193 \texttt{append}, 2 \texttt{merge}, 3,725 \texttt{replace}, 78 \texttt{transforms}, 1 \texttt{delete}, 1 \texttt{keep}, and 1 for \texttt{UNKNOWN} edits, as shown in Table \ref{tab:basic_edit_trans}, Table \ref{tab:edit_trans}, Table \ref{tab:suffix_trans_replace}, Table \ref{tab:suffix_trans_remove} and Table \ref{tab:suffix_trans_append}, in \ref{sec:appendix}. These edit operations, when applied to a source sentence, transform it into a corrected version. For \texttt{UNKNOWN} edits, we just copied the corrsponding source token as it is. The methodology for constructing these labels or classes is intricately detailed in Section \ref{sec:seqedit}.

\subsection{Implementation Details} 
\begin{itemize}
    \item \textbf{Encoders:} We used three prevalent standard pre-trained BERT-type models: BERT, RoBERTa, and XLNet. Specifically, we employed 'bert-base-uncased', 'roberta-base', and 'xlnet-base-cased' for our experiments. The models are available at the Hugging Face model repository \footnote{\url{https://huggingface.co/models}}.

\item \textbf{Training:} 
Following recent works in English GEC, we conduct experiments in the same setting with the restricted track of the BEA-19 GEC shared task \cite{Bryant2019}. Our training methodology was inspired by \cite{Omelianchuk2020}. A multi-stage training approach is adopted with the model first pretrained on the artificially generated pretraining dataset, Section \ref{sec:DAE}, until convergence. After pretraining, two more stages of finetuning are done. At each stage, we utilize the Adam optimizer with its default settings. An early stopping mechanism was implemented, terminating the training if no improvement was observed over 3 epochs, each consisting of 10K updates. The batch size was configured as 64 with accumulation size of 4 and learning rate of 1e-5. The number of epoches are 12, 6 and 1 for each traning stages, respectively. The hyperparameters were set using a combination of grid search and empirical tuning, based on the model's performance on the validation set. 

\item \textbf{Tweaking the inference:} 
In terms of inference tweaking, we adopted the strategy detailed in \cite{Omelianchuk2020}, applying confidence biases of 0.35, 0.23, and 0.17, alongside minimum error probability thresholds of 0.66, 0.61, and 0.63 for XLNet, RoBERTa, and BERT, respectively. These parameters were fine-tuned through grid search, ensuring the robustness and accuracy of our model's predictions.

\item \textbf{Newly introduced hyperparameters:}

In our proposed model, we have introduced two novel hyperparameters that are pivotal to the model's performance and functionality: the number of heads in the multi-head architecture and the lambda of the loss function.

\textbf{Number of Heads:}
\label{sec:heads}
The multi-head architecture is a core component of our model, enabling it to address different aspects of grammatical errors simultaneously. Each head is designed to focus on a specific subtask of GEC, such as Insertion, Deletion, Merge, Substitution, Transformation, Detection, or Correction. The number of heads is a critical hyperparameter that determines the model's capacity to handle the complexity and variety of grammatical errors. The number of heads is chosen using a trial-and-error approach, guided by our intuition about the primary edit types in GEC. We start with the most common edit operations: Delete, Replace, and Insert, which make three classification heads. In addition to the detection and correction, the total number of classification heads is five, Multi-head5. In our second experiment, we add two more edit operations that represent Transformation and Merging. The total number of classification heads is seven, Multi-head7.

\textbf{Lambda of the Loss Function:}
Lambda $\lambda$ in equation \eqref{eq:loss} represents a scaling factor that defines the magnitude of the contribution of each subtask’s loss to the total loss. Initially, we assigned a uniform value of $\lambda$ across all subtasks, determined through a random search over a range of values bewteen 0 and 1. Additoinally, we also explored different values of lambda for each subtask following a trial-and-error approach. However, this individualized adjustment led to a decrease in performance, possibly due to the intricate interplay and dependencies among the subtasks. We recognize that a more systematic and extensive exploration is needed, which we intend to pursue in future work.
\end{itemize}
\subsection{Evaluation}
To comprehensively assess the performance of our multi-head sequence tagging model, we employ two widely recognized evaluation metrics in the field of GEC: ERRANT, M2 Scorer and GLEU. These metrics enable us to evaluate the model's effectiveness in correcting grammatical errors and its ability to produce grammatically sound and fluent text.
\subsubsection{ERRANT}
The Error Annotation Toolkit (ERRANT) \cite{Bryant2017} is an annotation framework designed to automatically annotate grammatical errors in text, providing a standardized approach to error classification. It is the official evaluation metric for leading GEC benchmarks, including BEA-19. It's extensively used as the main evaluation metric for recent state-of-the-art GEC models, providing a standardized and comprehensive framework for assessing grammatical accuracy and the effectiveness of error correction.
\subsubsection{M2 Scorer}
The M2 Scorer \cite{dahlmeier-ng-2012-better} focuses on minimal edits and is renowned for its use in the CoNLL 2013 and 2014 shared tasks on GEC. It calculates the F0.5 score for word and phrase-level changes based on a lattice of edits made between the aligned original and corrected sentences. The M2 Scorer is particularly effective in evaluating the precision of grammatical corrections, making it a standard metric in the GEC field..
\subsubsection{GLEU} 
GLEU \cite{napoles-etal-2015-ground}, an adaptation of the BLEU metric \cite{papineni_bleu_2002} used in Machine Translation, measures the n-gram overlap between the system's output and human-corrected sentences. This metric is sensitive to fluency since it penalizes n-grams that have been changed in human corrections but are left unchanged by the system. Unlike metrics that require alignment between the original and corrected sentences, GLEU assesses fluency and minimal edits directly, providing a more holistic view of a system's ability to produce natural and well-formed text. 
\subsection{Experimental results}
The performance of our Multi-head7 models, both single and ensemble, is presented in Table \ref{tab:overall} alongside a selection of well-recognized GEC models. Our models demonstrate a notable improvement in the F0.5 score, significantly outperforming many of the compared models on both BEA-19 and CoNLL-14 test sets. Specifically, our single Multi-head7 (XLNet) model achieves an F0.5 score of 68.6 on CoNLL-14 and 74.4 on BEA-19. The ensemble version, Multi-head7 (XLNet+RoBERTa), further enhances these results, achieving an F0.5 score of 69.1 on CoNLL-14 and 77.0 on BEA-19, showcasing the strength of our model's architecture in handling grammatical errors. Although the 4×RPN ensemble \cite{didenko_redpennet_2023} shows a slightly higher score on BEA-19, it's essential to consider the complexity and computational resources involved in such ensemble of four larg models models and the fact that their approach is a combination of an autoregressive sequence-to-sequence and sequence tagging techniques, as discussed in Section \ref{sec:related}. The result on JFLEG dataset shows that our model achieves superior scores compared to all the baselines used in this work, except \cite{Stahlberg2021}, which may retain to the fact their model is a seq2seq model and the main contribution of \cite{Stahlberg2021} is introducing a huge synthetic pretraining dataset used to train Transformers. We think that most of the performance gain has come from this aspect. The result on JFLEG dataset will be discussed further in Section \ref{sec:jeleg}.
\begin{table*}[t]
	\centering
	\caption{Performance in English GEC benchmarks (i.e., CoNLL-14 and BEA-19 test). The single model scores are in the upper part, while the lower part shows the scores for the ensemble models.}
        \begin{adjustbox}{width=1\textwidth}
	\small
		\begin{tabular}{l|rrr|rrr|r}
			\hline
			& \multicolumn{3}{c}{\textbf{CoNLL-14(M2 Score)}} & \multicolumn{3}{c}{\textbf{BEA-19 test (ERRANT)}} & \textbf{JFLEG} \\
			\textbf{Method}  &  \textbf{P}  &  \textbf{R}  &  \textbf{F0.5}  &   \textbf{P}  &  \textbf{R}  &  \textbf{F0.5} & \textbf{$GLEU^+$}\\
			\hline
			gT5 xxl \cite{Rothe2021} & - & - & 65.7 & - & - & 69.8  & -\\
			gT5 base \cite{Rothe2021}  & - & - & 54.1 & - & - & 60.2 & -\\
			PIE \cite{Awasthi2019}  & 66.1 & 43.0 & 59.7 & - & - & - & 60.3 \\
			Stahlberg 2021 \cite{Stahlberg2021}    & 72.8 & 49.5 & 66.6 & 72.1 & 64.4 & 70.4 & \textbf{64.7} \\
			GECTor \cite{Omelianchuk2020} & 77.5 & 40.0 & 61.3 & 79.2 & 53.9 & 72.4 & - \\
                RPN(MPRBASE) \cite{didenko_redpennet_2023} & - & - & - & 80.8 & 56.7 & 74.4 & - \\	
                roberta-base-scorer-only \cite{sorokin_improved_2022} & 72.6 & 39.5 & 63.9 & 82.8 & 52.4 & 74.2& - \\	
   \hline
			\textbf{Ours, Multi-head7 (XLNet)} & 74.4	 & 52.3 & \textbf{68.6} & 82.6 & 53.2 & \textbf{74.4} & 61.6 \\
			\hline
			PIE \cite{Awasthi2019}  & 68.3 & 43.2 & 61.2 & - & - & - & 61.0\\
			Stahlberg 2021 \cite{Stahlberg2021}  & 75.6 & 49.3 & 68.3 & 77.7 & \textbf{65.4} & 74.9 & \textbf{64.7}\\
			GECTor \cite{Omelianchuk2020} & 78.2 & 41.5 & 66.5 & 79.4 & 57.2 & 73.7 &  \\
   			4×RPN ensemble \cite{didenko_redpennet_2023} & - & - & - & 86.6 & 54.8 & \textbf{77.6} &  \\
			\hline
			\textbf{Ours, Multi-head7(XLNet+RoBERTa)}  & 75.58 & 51.32 & \textbf{69.1} & 85.7 & 54.8 & 77.0 & 61.7 \\
			\hline
		\end{tabular}
	\label{tab:overall}
 \end{adjustbox}
\end{table*}
\subsection{Ablation study on the Model Architecture}
To study the effect of different experimental settings, we design several experiments as follows:
\subsubsection{Number of classification heads}
As shown in the upper part of Table \ref{tab:numberofheads}, several architectures with a different number of classification heads are evaluated. First, the base model (Base Model XLNet) in Table \ref{tab:numberofheads} is our implementation of GECToR model as described in Section \ref{sec:baseArch}. The reported result is after three stages of training using the same training data and settings as our other architectures. From the experimental results in Table  \ref{tab:numberofheads}, the Multi-head5 model, mentioned at \ref{sec:heads}, outperforms the base model. Additionally, we can observe an improvement after adding two heads, Multi-head7 model. In conclusion, the seven-heads architecture, Multi-head7, gives the best result, which might return to the leveraging of the multi-task learning, especially when the tasks are related to the main task, as in this case. Using the multi-head architecture helps the model's learning to distinguish between different types of edits that convert the source sentence to a corrected one. It is worth noting that adding the classification heads doesn’t change the inference speed of the model compared to the base model, since we are using the multi-head architecture in the training stage only. 

The observed performance improvement with an increased number of heads from 5 to 7 underscores the multi-head architecture's effectiveness in handling distinct grammatical correction tasks. Each additional head allows for more nuanced error handling, contributing to the model's overall performance. However, it's crucial to consider the balance between complexity and benefit. While adding more heads could theoretically improve performance by addressing more specific subtasks, it also increases the model's complexity and computational demand. Moreover, the marginal gains might diminish as the number of heads grows, and the risk of overfitting escalates if not managed with sufficient data and regularization. Future exploration in this area might focus on identifying the optimal number of heads that provide the best performance without unduly complicating the model's architecture.
\begin{table*}[t]
	\caption{The performance of XLNet-based model on BEA-19 dev: upper part includes the multi-head with different number of heads. The effect of the transformer encoders is in the middle part. The lower part shows the effect of the proposed Seq2Edit Function. All results are obtained after the three stages of training.}
	\begin{center}
		\begin{tabular}{lrrrrrr}
			\hline
			Method  &   P  &  R  &  F0.5 \\
			\hline
			Base model(XLNet) & 66.63   & 35.54   & 56.71 \\
			Multi-head5(XLNet) & 67.22   & 35.89   & 57.23 \\
			Multi-head7(XLNet) & \textbf{68.67}   & \textbf{36.34}   & \textbf{58.29} \\
			\hline
			Multi-head7 (BERT) & 66.45   & 33.79   & 55.68 \\
			Multi-head7 (RoBERTa) & 67.11   & \textbf{36.93}   & 57.68 \\ 
			Multi-head7 (XLNet) & \textbf{68.67}   & 36.34   & \textbf{58.29} \\
			\hline
  		\textbf{The Effect of our Seq2Edit Function} \\
            GECToR model(XLNet) with their Edit2Seq & 66.0 & 33.8 & 55.5 \\
            GECToR model(XLNet) with our Edit2Seq & \textbf{66.6}   &  \textbf{35.5}   &  \textbf{56.7} \\
            			\hline

		\end{tabular}
	\end{center}
	\label{tab:numberofheads}
\end{table*}
\subsubsection{Pretrained Encoder Type}
To test the effect of different BERT-like models as encoders,  we fine-tuned BERT \cite{Devlin2019}, RoBERTa \cite{Liu2019}, XLNet \cite{Yang2019} with the same hyperparameters setup. The results in the middle part of Table \ref{tab:numberofheads} show that RoBERTa-based model did better than the BERT-based one, since RoBERTa robustly optimized the BERT approach by retraining it with improved training methodology \cite{Liu2019}. Moreover, the XLNet-based model has achieved the best overall result. That may return to the nature of XLNet that uses permutation-based training, which can handle dependencies well. In addition, Transformer XL was used as the base architecture, which showed good performance on different tasks \cite{Yang2019}.
\subsubsection{The effect of Denoising auto-encoder}
As shown in Table \ref{tab:numberofheads}, using the pretraining synthetic dataset generated with our denoising auto-encoder has improved the performace of the model compared to the dataset used in GECToR model \cite{Omelianchuk2020}. That improvement returns to the fact that the distribution of the artificially generated errors is better than the one generated by the previous work and used by GECToR model  .  
\subsubsection{Pre-training and Finetuning stages}
The result at the upper part of Table \ref{tab:pretraining} demonstrate the effectiveness of the three-step training strategy. We can observe an improvement when the model is pretrained on the synthetic dataset mentioned in Section \ref{sec:DAE}.  The result also shows that finetuning the model for one epoch on BEA (W\&I) \cite{Bryant2019} has improved the performance further. This proves that pre-training gives the model much better initial parameters. On the other hand, the finetuning stage has helped in domain adaptation. Moreover, enriching the pretraining dataset with more errors of specific type, mentioned in Section \ref{sec:DAE}, has improved the performance further compared to the one of GECTOR \cite{Omelianchuk2020}, as shown at the lower part of Table \ref{tab:pretraining}. 
\begin{table*}[t]
	\caption{The effect of the training stages and the pretraining dataset for Multi-head7(XLNet) and our implementation of GECToR(XLNet) \cite{Omelianchuk2020} with  the proposed Seq2Edit function evaluated on BEA-19 dev .}
	\small
	\begin{center}
		\begin{tabular}{lrrrrrr}
			\hline
			\textbf{Method}  &   \textbf{P}  &  \textbf{R}  &  \textbf{F0.5} \\
			\hline
			Our Pretraining data only  & 59.96 &  20.77 & 43.54 \\
			Training only & 62.68 & 31.79 & 52.48 \\
			Training+finetuning & 64.39 & 34.81 & 55.04 \\
			Our Pretraining data+ training+finetuning & 68.67   & \textbf{36.34}   & \textbf{58.29} \\
			GECToR pretraining  data+training+finetuning & \textbf{68.71}   & 35.20   & 57.72 \\
   			\hline
      			\textbf{GECToR(XLNet) (our implementation)} \\
            GECToR model with our Our Pretraining data+ training+finetuning & \textbf{66.63} & 35.54 & \textbf{56.71} \\
            GECToR model with GECTor Pretraining data+ training+finetuning & 58.36	& 44.52	& 54.95  \\
            			\hline                                                
		\end{tabular}
	\end{center}
	\label{tab:pretraining}
\end{table*}
\begin{table}[t]
	\caption{Performance of the Multi-head7 XLNet-based model using different lambda on BEA-19 dev. }
	\small
 	\begin{center}
		\begin{tabular}{lrrrr}
			\hline
			\textbf{lambda} $\lambda$ &\textbf{1}&\textbf{0.5}&\textbf{0.3}&\textbf{0.25} \\
			\hline
			P & \textbf{68.77} & 68.67  & 68.55  & 67.30 \\
			R & 35.16  & \textbf{36.34}  & 35.76  & 36.21  \\
			F0.5 & 57.73 & \textbf{58.29} &  57.93 & 57.44\\
			\hline
		\end{tabular}
  \end{center}
	\label{tab:lambda}
\end{table} 
\subsubsection{Lambda of the loss function}
Lambda $\lambda$ in equation \eqref{eq:loss} represents a scaling factor that defines the magnitude of the contribution of each subtask’s loss to the total loss. As shown in Table \ref{tab:lambda}, we found that the best value is ($\lambda=0.5$) chosen by a random search over several values. That was expected since we need to give more attention to the main task while leveraging the information coming from related tasks. It is worth noting that we used the same value of $\lambda$, in equation \eqref{eq:loss}, for all the classification heads except the correction one. Different values can be assigned for each head, but we left that for future work.
\subsubsection{The Effect of our Edit2Seq Function}
In Table \ref{tab:numberofheads}, we present a comparative analysis of the performance of our implementation of the GECToR model \cite{Omelianchuk2020}. This comparison contrasts the results obtained using the original GECToR's Seq2Edit function with those achieved utilizing our modified Seq2Edit function, as detailed in Section \ref{sec:seqedit}. The data indicate that incorporating our Seq2Edit function enhances the model's performance, yielding an improvement of over 1 point in the $F_{0.5}$ score. This significant enhancement is largely attributable to the superior vocabulary coverage afforded by our refined Seq2Edit approach.
\subsubsection{On JFLEG dataset for fluency evaluation}
\label{sec:jeleg}
To align our evaluation with related work, we present span-based ERRANT F0.5 scores on the development and test sets (BEA-dev and BEA-test) of the BEA-2019 shared task \cite{Bryant2019}. Additionally, we have incorporated the GLEU metric to provide a comprehensive evaluation of our model's fluency. This ensures a multifaceted analysis of the text output quality. The JFLEG test set \cite{napoles-etal-2017-jfleg} is particularly notable as it doesn't limit corrections to minimal error spans and doesn't code the errors; instead, it involves holistic sentence rewrites.

Our model was applied to the JFLEG dataset without prior spellchecking or preprocessing, achieving GLEU scores of 61.62 and 61.72 for the single and ensemble models, respectively, as detailed in Table ~\ref{tab:overall}. This performance surpasses that of the PIE model. However, as anticipated, the GLEU scores are modest compared to the results from the sequence-to-sequence model of \cite{Stahlberg2021}. We attribute this to the inherent characteristics of sequence tagging models, particularly their parallel predictions and the inherent challenge in making extensive modifications (e.g., multi-token edits). This limitation is a recognized aspect of sequence tagging models. Furthermore, it's important to note that our model was fine-tuned on the BEA-2019 dataset, not the JFLEG development set, which may also influence the evaluation outcomes.

\section{Limitations and Future Directions}
Our proposed approach, which employs Seq2Edit models, shows promising results. However, there are inherent limitations when compared to traditional Seq2Seq models. First, our models, though effective on datasets with minimal edits, struggle with texts requiring extensive or drastic changes. Second, 
we employed standard BERT-base variants due to their balance between performance and computational efficiency, enabling us to focus on the main contribution of our multi-head model. We are planning to explore more advanced and recent models in future work, as computational resources permit. Furthermore, we want to delve deeper into the concept of applying the proposed approach to languages other than English and validate its feasibility.

\section{Conclusion}
We propose a simple yet novel approach to improve the efficiency of GEC by dividing the GEC task into seven related subtasks. We use an efficient sequence tagging model to identify the text edits needed to transform an incorrect source sentence into a grammatically correct one. Combined with a new way of generating pretraining data and a careful design of the Seq2Edit function, our approach performs comparably to the state-of-the-art models. To the best of our knowledge, we are the first to use multi-head architecture for GEC in sequence tagging settings. The seven used edit operations are arguably more similar to how human writes or edits text. 
\bibliographystyle{unsrt} 
\bibliography{Main.bib}

\begin{thebibliography}{10}

\bibitem{Jia2013}
Zhongye Jia, Peilu Wang, and Hai Zhao.
\newblock Grammatical error correction as multiclass classification with single model.
\newblock In {\em Proceedings of the Seventeenth Conference on Computational Natural Language Learning: Shared Task}, pages 74--81, Sofia, Bulgaria, August 2013. Association for Computational Linguistics.

\bibitem{Rothe2021}
Sascha Rothe, Jonathan Mallinson, Eric Malmi, Sebastian Krause, and Aliaksei Severyn.
\newblock A simple recipe for multilingual grammatical error correction.
\newblock {\em arXiv preprint arXiv:2106.03830}, 2021.

\bibitem{Bryant2019}
Christopher Bryant, Mariano Felice, {\O}istein~E. Andersen, and Ted Briscoe.
\newblock The {BEA}-2019 shared task on grammatical error correction.
\newblock In {\em Proceedings of the Fourteenth Workshop on Innovative Use of NLP for Building Educational Applications}, pages 52--75, Florence, Italy, August 2019. Association for Computational Linguistics.

\bibitem{Kiyono2019}
Shun Kiyono, Jun Suzuki, Masato Mita, Tomoya Mizumoto, and Kentaro Inui.
\newblock An empirical study of incorporating pseudo data into grammatical error correction.
\newblock In {\em Proceedings of the 2019 Conference on Empirical Methods in Natural Language Processing and the 9th International Joint Conference on Natural Language Processing (EMNLP-IJCNLP)}, pages 1236--1242, Hong Kong, China, November 2019. Association for Computational Linguistics.

\bibitem{Zhao2019}
Wei Zhao, Liang Wang, Kewei Shen, Ruoyu Jia, and Jingming Liu.
\newblock Improving grammatical error correction via pre-training a copy-augmented architecture with unlabeled data.
\newblock In {\em Proceedings of the 2019 Conference of the North {A}merican Chapter of the Association for Computational Linguistics: Human Language Technologies, Volume 1 (Long and Short Papers)}, pages 156--165, Minneapolis, Minnesota, June 2019. Association for Computational Linguistics.

\bibitem{Stahlberg2021}
Felix Stahlberg and Shankar Kumar.
\newblock Synthetic data generation for grammatical error correction with tagged corruption models.
\newblock In {\em Proceedings of the 16th Workshop on Innovative Use of NLP for Building Educational Applications}, pages 37--47, Online, April 2021. Association for Computational Linguistics.

\bibitem{Lichtarge2019}
Jared Lichtarge, Chris Alberti, Shankar Kumar, Noam Shazeer, Niki Parmar, and Simon Tong.
\newblock Corpora generation for grammatical error correction.
\newblock In {\em Proceedings of the 2019 Conference of the North {A}merican Chapter of the Association for Computational Linguistics: Human Language Technologies, Volume 1 (Long and Short Papers)}, pages 3291--3301, Minneapolis, Minnesota, June 2019. Association for Computational Linguistics.

\bibitem{al2018bidirectional}
Kamal Al-Sabahi, Zhang Zuping, and Yang Kang.
\newblock Bidirectional attentional encoder-decoder model and bidirectional beam search for abstractive summarization.
\newblock {\em arXiv preprint arXiv:1809.06662}, 2018.

\bibitem{Awasthi2019}
Abhijeet Awasthi, Sunita Sarawagi, Rasna Goyal, Sabyasachi Ghosh, and Vihari Piratla.
\newblock Parallel iterative edit models for local sequence transduction.
\newblock In {\em Proceedings of the 2019 Conference on Empirical Methods in Natural Language Processing and the 9th International Joint Conference on Natural Language Processing (EMNLP-IJCNLP)}, pages 4260--4270, Hong Kong, China, November 2019. Association for Computational Linguistics.

\bibitem{Omelianchuk2020}
Kostiantyn Omelianchuk, Vitaliy Atrasevych, Artem Chernodub, and Oleksandr Skurzhanskyi.
\newblock {GECT}o{R} {--} grammatical error correction: Tag, not rewrite.
\newblock In {\em Proceedings of the Fifteenth Workshop on Innovative Use of NLP for Building Educational Applications}, pages 163--170, Seattle, WA, USA → Online, July 2020. Association for Computational Linguistics.

\bibitem{Li2022}
Jiquan Li, Junliang Guo, Yongxin Zhu, Xin Sheng, Deqiang Jiang, Bo~Ren, and Linli Xu.
\newblock Sequence-to-action: Grammatical error correction with action guided sequence generation.
\newblock {\em Proceedings of the AAAI Conference on Artificial Intelligence}, 36(10):10974--10982, Jun. 2022.

\bibitem{Kamal2018}
Kamal Al-Sabahi, Zhang Zuping, and Mohammed Nadher.
\newblock A hierarchical structured self-attentive model for extractive document summarization (hssas).
\newblock {\em IEEE Access}, 6:24205--24212, 2018.

\bibitem{mohsen2020hierarchical}
Farida Mohsen, Jiayang Wang, and Kamal Al-Sabahi.
\newblock A hierarchical self-attentive neural extractive summarizer via reinforcement learning (hsasrl).
\newblock {\em Applied Intelligence}, 50(9):2633--2646, 2020.

\bibitem{Changpinyo2018}
Soravit Changpinyo, Hexiang Hu, and Fei Sha.
\newblock Multi-task learning for sequence tagging: An empirical study.
\newblock {\em arXiv preprint arXiv:1808.04151}, 2018.

\bibitem{10181308}
Kamal Al-Sabahi and Kang Yang.
\newblock Supervised copy mechanism for grammatical error correction.
\newblock {\em IEEE Access}, 11:72374--72383, 2023.

\bibitem{Schröder2020}
Fynn Schr{\"o}der and Chris Biemann.
\newblock Estimating the influence of auxiliary tasks for multi-task learning of sequence tagging tasks.
\newblock In {\em Proceedings of the 58th Annual Meeting of the Association for Computational Linguistics}, pages 2971--2985, Online, July 2020. Association for Computational Linguistics.

\bibitem{Wang2021}
Haoxiang Wang, Han Zhao, and Bo~Li.
\newblock Bridging multi-task learning and meta-learning: Towards efficient training and effective adaptation.
\newblock {\em arXiv preprint arXiv:2106.09017}, 2021.

\bibitem{Crawshaw2020}
Michael Crawshaw.
\newblock Multi-task learning with deep neural networks: A survey.
\newblock {\em arXiv preprint arXiv:2009.09796}, 2020.

\bibitem{Ruder2017}
Sebastian Ruder.
\newblock An overview of multi-task learning in deep neural networks.
\newblock {\em arXiv preprint arXiv:1706.05098}, 2017.

\bibitem{Zhang2021}
Y~Zhang and Q~Yang.
\newblock A survey on multi-task learning.
\newblock {\em IEEE Transactions on Knowledge and Data Engineering}, page~1, 2021.

\bibitem{Fynn2020}
Fynn Schr{\"o}der and Chris Biemann.
\newblock Estimating the influence of auxiliary tasks for multi-task learning of sequence tagging tasks.
\newblock In {\em Proceedings of the 58th Annual Meeting of the Association for Computational Linguistics}, pages 2971--2985, Online, July 2020. Association for Computational Linguistics.

\bibitem{Lichtarge2020}
Jared Lichtarge, Chris Alberti, and Shankar Kumar.
\newblock Data weighted training strategies for grammatical error correction.
\newblock {\em Transactions of the Association for Computational Linguistics}, 8:634--646, 2020.

\bibitem{didenko_redpennet_2023}
Bohdan Didenko and Andrii Sameliuk.
\newblock {RedPenNet} for {Grammatical} {Error} {Correction}: {Outputs} to {Tokens}, {Attentions} to {Spans}.
\newblock In Mariana Romanyshyn, editor, {\em Proceedings of the {Second} {Ukrainian} {Natural} {Language} {Processing} {Workshop} ({UNLP})}, pages 121--131, Dubrovnik, Croatia, May 2023. Association for Computational Linguistics.

\bibitem{sorokin_improved_2022}
Alexey Sorokin.
\newblock Improved grammatical error correction by ranking elementary edits.
\newblock In Yoav Goldberg, Zornitsa Kozareva, and Yue Zhang, editors, {\em Proceedings of the 2022 {Conference} on {Empirical} {Methods} in {Natural} {Language} {Processing}}, pages 11416--11429, Abu Dhabi, United Arab Emirates, December 2022. Association for Computational Linguistics.

\bibitem{Xue2021}
Linting Xue, Noah Constant, Adam Roberts, Mihir Kale, Rami Al-Rfou, Aditya Siddhant, Aditya Barua, and Colin Raffel.
\newblock m{T}5: A massively multilingual pre-trained text-to-text transformer.
\newblock In {\em Proceedings of the 2021 Conference of the North American Chapter of the Association for Computational Linguistics: Human Language Technologies}, pages 483--498, Online, June 2021. Association for Computational Linguistics.

\bibitem{Bryant2017}
Christopher Bryant, Mariano Felice, and Ted Briscoe.
\newblock Automatic annotation and evaluation of error types for grammatical error correction.
\newblock In {\em Proceedings of the 55th Annual Meeting of the Association for Computational Linguistics (Volume 1: Long Papers)}, pages 793--805, Vancouver, Canada, July 2017. Association for Computational Linguistics.

\bibitem{Wan2020}
Zhaohong Wan, Xiaojun Wan, and Wenguang Wang.
\newblock Improving grammatical error correction with data augmentation by editing latent representation.
\newblock In {\em Proceedings of the 28th International Conference on Computational Linguistics}, pages 2202--2212, Barcelona, Spain (Online), December 2020. International Committee on Computational Linguistics.

\bibitem{Chelba2013}
Ciprian Chelba, Tomas Mikolov, Mike Schuster, Qi~Ge, Thorsten Brants, Phillipp Koehn, and Tony Robinson.
\newblock One billion word benchmark for measuring progress in statistical language modeling.
\newblock {\em arXiv preprint arXiv:1312.3005}, 2013.

\bibitem{KakaoBrain2019}
Yo~Joong Choe, Jiyeon Ham, Kyubyong Park, and Yeoil Yoon.
\newblock A neural grammatical error correction system built on better pre-training and sequential transfer learning.
\newblock In {\em Proceedings of the Fourteenth Workshop on Innovative Use of NLP for Building Educational Applications}, pages 213--227, Florence, Italy, August 2019. Association for Computational Linguistics.

\bibitem{Vaswani2017}
Ashish Vaswani, Noam Shazeer, Niki Parmar, Jakob Uszkoreit, Llion Jones, Aidan~N Gomez, Łukasz Kaiser, and Illia Polosukhin.
\newblock Attention is all you need.
\newblock {\em Advances in neural information processing systems}, 30, 2017.

\bibitem{Mizumoto2011}
Tomoya Mizumoto, Mamoru Komachi, Masaaki Nagata, and Yuji Matsumoto.
\newblock Mining revision log of language learning {SNS} for automated {J}apanese error correction of second language learners.
\newblock In {\em Proceedings of 5th International Joint Conference on Natural Language Processing}, pages 147--155, Chiang Mai, Thailand, November 2011. Asian Federation of Natural Language Processing.

\bibitem{Dahlmeier2013}
Daniel Dahlmeier, Hwee~Tou Ng, and Siew~Mei Wu.
\newblock Building a large annotated corpus of learner {E}nglish: The {NUS} corpus of learner {E}nglish.
\newblock In {\em Proceedings of the Eighth Workshop on Innovative Use of {NLP} for Building Educational Applications}, pages 22--31, Atlanta, Georgia, June 2013. Association for Computational Linguistics.

\bibitem{Yannakoudakis2011}
Helen Yannakoudakis, Ted Briscoe, and Ben Medlock.
\newblock A new dataset and method for automatically grading {ESOL} texts.
\newblock In {\em Proceedings of the 49th Annual Meeting of the Association for Computational Linguistics: Human Language Technologies}, pages 180--189, Portland, Oregon, USA, June 2011. Association for Computational Linguistics.

\bibitem{Ng2014}
Hwee~Tou Ng, Siew~Mei Wu, Ted Briscoe, Christian Hadiwinoto, Raymond~Hendy Susanto, and Christopher Bryant.
\newblock The {C}o{NLL}-2014 shared task on grammatical error correction.
\newblock In {\em Proceedings of the Eighteenth Conference on Computational Natural Language Learning: Shared Task}, pages 1--14, Baltimore, Maryland, June 2014. Association for Computational Linguistics.

\bibitem{napoles-etal-2017-jfleg}
Courtney Napoles, Keisuke Sakaguchi, and Joel Tetreault.
\newblock {JFLEG}: A fluency corpus and benchmark for grammatical error correction.
\newblock In {\em Proceedings of the 15th Conference of the {E}uropean Chapter of the Association for Computational Linguistics: Volume 2, Short Papers}, pages 229--234, Valencia, Spain, April 2017. Association for Computational Linguistics.

\bibitem{Ng2013}
Hwee~Tou Ng, Siew~Mei Wu, Yuanbin Wu, Christian Hadiwinoto, and Joel Tetreault.
\newblock The {C}o{NLL}-2013 shared task on grammatical error correction.
\newblock In {\em Proceedings of the Seventeenth Conference on Computational Natural Language Learning: Shared Task}, pages 1--12, Sofia, Bulgaria, August 2013. Association for Computational Linguistics.

\bibitem{dahlmeier-ng-2012-better}
Daniel Dahlmeier and Hwee~Tou Ng.
\newblock Better evaluation for grammatical error correction.
\newblock In Eric Fosler-Lussier, Ellen Riloff, and Srinivas Bangalore, editors, {\em Proceedings of the 2012 Conference of the North {A}merican Chapter of the Association for Computational Linguistics: Human Language Technologies}, pages 568--572, Montr{\'e}al, Canada, June 2012. Association for Computational Linguistics.

\bibitem{napoles-etal-2015-ground}
Courtney Napoles, Keisuke Sakaguchi, Matt Post, and Joel Tetreault.
\newblock Ground truth for grammatical error correction metrics.
\newblock In Chengqing Zong and Michael Strube, editors, {\em Proceedings of the 53rd Annual Meeting of the Association for Computational Linguistics and the 7th International Joint Conference on Natural Language Processing (Volume 2: Short Papers)}, pages 588--593, Beijing, China, July 2015. Association for Computational Linguistics.

\bibitem{papineni_bleu_2002}
Kishore Papineni, Salim Roukos, Todd Ward, and Wei-Jing Zhu.
\newblock Bleu: a {Method} for {Automatic} {Evaluation} of {Machine} {Translation}.
\newblock In Pierre Isabelle, Eugene Charniak, and Dekang Lin, editors, {\em Proceedings of the 40th {Annual} {Meeting} of the {Association} for {Computational} {Linguistics}}, pages 311--318, Philadelphia, Pennsylvania, USA, July 2002. Association for Computational Linguistics.

\bibitem{Devlin2019}
Jacob Devlin, Ming-Wei Chang, Kenton Lee, and Kristina Toutanova.
\newblock {BERT}: Pre-training of deep bidirectional transformers for language understanding.
\newblock In {\em Proceedings of the 2019 Conference of the North {A}merican Chapter of the Association for Computational Linguistics: Human Language Technologies, Volume 1 (Long and Short Papers)}, pages 4171--4186, Minneapolis, Minnesota, June 2019. Association for Computational Linguistics.

\bibitem{Liu2019}
Yinhan Liu, Myle Ott, Naman Goyal, Jingfei Du, Mandar Joshi, Danqi Chen, Omer Levy, Mike Lewis, Luke Zettlemoyer, and Veselin Stoyanov.
\newblock Roberta: A robustly optimized bert pretraining approach.
\newblock {\em arXiv preprint arXiv:1907.11692}, 2019.

\bibitem{Yang2019}
Zhilin Yang, Zihang Dai, Yiming Yang, Jaime Carbonell, Russ~R Salakhutdinov, and Quoc~V Le.
\newblock Xlnet: Generalized autoregressive pretraining for language understanding.
\newblock {\em Advances in neural information processing systems}, 32, 2019.

\end{thebibliography}

\appendix
\section{Appendix}
\label{sec:appendix}
\begin{table*}
	\caption{List of basic transformations.}
	\small
	\begin{center}
		\begin{tabular}{llll}
			\hline
			id & Basic Transformation & Transformation suffix & Tag \\ 
			\hline
			1 & KEEP &  & \$KEEP  \\ 
			2 & DELETE &  & \$DELETE \\ 
			3 & APPEND & a & \$APPEND\_a \\ 
			… & … & … & … \\ 
			973 & APPEND & yourself & \$APPEND\_yourself \\ 
			974 & REPLACE & ! & \$REPLACE\_! \\ 
			… & … & … & … \\ 
			4902 & REPLACE & yourselves & \$REPLACE\_yourselves \\ 
			4903 & MERGE & HYPHEN & \$MERGE\_HYPHEN \\ 
			4904 & MERGE & SPACE & \$MERGE\_SPACE \\ 
		\end{tabular}
	\end{center}
	\label{tab:basic_edit_trans}
\end{table*}

\begin{table*}
	\caption{List of other transformations.}
	\small
	\begin{center}
		\begin{tabular}{llll}
			\hline
			id  & Transformation suffix & Tag \\ 
			\hline
			4975 & AGREEMENT\_PLURAL & \$TRANSFORM\_AGREEMENT\_PLURAL \\ 
			4976 & AGREEMENT\_SINGULAR & \$TRANSFORM\_AGREEMENT\_SINGULAR \\ 
			4977 & CASE\_CAPITAL & \$TRANSFORM\_CASE\_CAPITAL \\ 
			4978 & CASE\_LOWER & \$TRANSFORM\_CASE\_LOWER \\ 
			4979  & CASE\_UPPER & \$TRANSFORM\_CASE\_UPPER \\ 
			4980  & SPLIT\_HYPHEN & \$TRANSFORM\_SPLIT\_HYPHEN \\ 
			4981 & VERB\_VBD\_VB & \$TRANSFORM\_VERB\_VBD\_VB \\ 
			4982 & VERB\_VBD\_VBG & \$TRANSFORM\_VERB\_VBD\_VBG \\ 
			4983 & VERB\_VBD\_VBN & \$TRANSFORM\_VERB\_VBD\_VBN \\ 
			4984 & VERB\_VBD\_VBZ & \$TRANSFORM\_VERB\_VBD\_VBZ \\  
			4985 & VERB\_VBG\_VB & \$TRANSFORM\_VERB\_VBG\_VB \\ 
			4986 & VERB\_VBG\_VBD & \$TRANSFORM\_VERB\_VBG\_VBD \\ 
			4987 & VERB\_VBG\_VBN & \$TRANSFORM\_VERB\_VBG\_VBN \\ 
			4988 & VERB\_VBG\_VBZ & \$TRANSFORM\_VERB\_VBG\_VBZ \\ 
			4989 & VERB\_VBN\_VB & \$TRANSFORM\_VERB\_VBN\_VB \\   
			4990 & VERB\_VBN\_VBD & \$TRANSFORM\_VERB\_VBN\_VBD \\ 
			4991 & VERB\_VBN\_VBG & \$TRANSFORM\_VERB\_VBN\_VBG \\ 
			4992 & VERB\_VBN\_VBZ & \$TRANSFORM\_VERB\_VBN\_VBZ \\ 
			4993 & VERB\_VBZ\_VB & \$TRANSFORM\_VERB\_VBZ\_VB \\ 
			4994 & VERB\_VBZ\_VBD & \$TRANSFORM\_VERB\_VBZ\_VBD \\ 
			4995 & VERB\_VBZ\_VBG & \$TRANSFORM\_VERB\_VBZ\_VBG \\ 
			4996 & VERB\_VBZ\_VBN & \$TRANSFORM\_VERB\_VBZ\_VBN \\ 
			4997 & VERB\_VB\_VBD & \$TRANSFORM\_VERB\_VB\_VBD \\ 
			4998 & VERB\_VB\_VBG & \$TRANSFORM\_VERB\_VB\_VBG \\ 
			4999 & VERB\_VB\_VBN & \$TRANSFORM\_VERB\_VB\_VBN \\ 
			5000 & VERB\_VB\_VBZ & \$TRANSFORM\_VERB\_VB\_VBZ \\ 
			
		\end{tabular}
	\end{center}
	\label{tab:edit_trans}
\end{table*}

\begin{table*}
	\caption{List of suffix replace transformations.}
 	\small
	\begin{center}
		\begin{tabular}{ll ll }
			\hline
			id  & Transformation suffix & Tag \\ 
			\hline
			4905  & AL\_TO\_E & \$SUFFIXTRANSFORM\_AL\_TO\_E \\ 
			4925  & ATION\_TO\_ING & \$SUFFIXTRANSFORM\_ATION\_TO\_ING \\ 
			4926  & CE\_TO\_T & \$SUFFIXTRANSFORM\_CE\_TO\_T  \\ 
			4927  & D\_TO\_S & \$SUFFIXTRANSFORM\_D\_TO\_S \\ 
			4928  & D\_TO\_T & \$SUFFIXTRANSFORM\_D\_TO\_T \\ 
			4929  & ED\_TO\_ING & \$SUFFIXTRANSFORM\_ED\_TO\_ING \\ 
			4930  & ED\_TO\_S & \$SUFFIXTRANSFORM\_ED\_TO\_S \\ 
			4931  & ER\_TO\_EST & \$SUFFIXTRANSFORM\_ER\_TO\_EST \\ 
			4932  & EST\_TO\_ER & \$SUFFIXTRANSFORM\_EST\_TO\_ER \\ 
			4933  & E\_TO\_AL & \$SUFFIXTRANSFORM\_E\_TO\_AL \\ 
			4934  & E\_TO\_ING & \$SUFFIXTRANSFORM\_E\_TO\_ING \\ 
			4935  & ICAL\_TO\_Y & \$SUFFIXTRANSFORM\_ICAL\_TO\_Y \\ 
			4936  & IC\_TO\_Y & \$SUFFIXTRANSFORM\_IC\_TO\_Y \\ 
			4937  & IES\_TO\_Y & \$SUFFIXTRANSFORM\_IES\_TO\_Y \\ 
			4938  & ILY\_TO\_Y & \$SUFFIXTRANSFORM\_ILY\_TO\_Y \\ 
			4939  & ING\_TO\_ATION & \$SUFFIXTRANSFORM\_ING\_TO\_ATION \\ 
			4940  & ING\_TO\_E & \$SUFFIXTRANSFORM\_ING\_TO\_E \\ 
			4941  & ING\_TO\_ED & \$SUFFIXTRANSFORM\_ING\_TO\_ED \\ 
			4942  & ING\_TO\_ION & \$SUFFIXTRANSFORM\_ING\_TO\_ION \\ 
			4943  & ING\_TO\_S & \$SUFFIXTRANSFORM\_ING\_TO\_S \\ 
			4944  & ION\_TO\_ING & \$SUFFIXTRANSFORM\_ION\_TO\_ING \\ 
			4945  & N\_TO\_ING & \$SUFFIXTRANSFORM\_N\_TO\_ING \\ 
			4963  & S\_TO\_D & \$SUFFIXTRANSFORM\_S\_TO\_D \\ 
			4964  & S\_TO\_ED & \$SUFFIXTRANSFORM\_S\_TO\_ED \\ 
			4965  & S\_TO\_ING & \$SUFFIXTRANSFORM\_S\_TO\_ING \\ 
			4966  & S\_TO\_T & \$SUFFIXTRANSFORM\_S\_TO\_T \\ 
			4967  & T\_TO\_CE & \$SUFFIXTRANSFORM\_T\_TO\_CE \\ 
			4968  & T\_TO\_D & \$SUFFIXTRANSFORM\_T\_TO\_D \\ 
			4969  & T\_TO\_S & \$SUFFIXTRANSFORM\_T\_TO\_S \\ 
			4970  & Y\_TO\_IC & \$SUFFIXTRANSFORM\_Y\_TO\_IC \\ 
			4971  & Y\_TO\_ICAL & \$SUFFIXTRANSFORM\_Y\_TO\_ICAL \\ 
			4972  & Y\_TO\_IED & \$SUFFIXTRANSFORM\_Y\_TO\_IED \\ 
			4973  & Y\_TO\_IES & \$SUFFIXTRANSFORM\_Y\_TO\_IES \\ 
			4974  & Y\_TO\_ILY & \$SUFFIXTRANSFORM\_Y\_TO\_ILY \\ 
						\hline
		\end{tabular}
	\end{center}
	\label{tab:suffix_trans_replace}
\end{table*}

\begin{table*}
	\caption{List of suffix remove transformations.}
 	\small
	\begin{center}
		\begin{tabular}{ll l l }
			\hline
			id  & Transformation suffix & Tag \\ 
			\hline 
			4946  & REMOVE\_able & \$SUFFIXTRANSFORM\_REMOVE\_able \\ 
			4947  & REMOVE\_age & \$SUFFIXTRANSFORM\_REMOVE\_age \\ 
			4948  & REMOVE\_al & \$SUFFIXTRANSFORM\_REMOVE\_al \\ 
			4949  & REMOVE\_ation & \$SUFFIXTRANSFORM\_REMOVE\_ation \\ 
			4950  & REMOVE\_d & \$SUFFIXTRANSFORM\_REMOVE\_d \\ 
			4951  & REMOVE\_ed & \$SUFFIXTRANSFORM\_REMOVE\_ed \\ 
			4952  & REMOVE\_er & \$SUFFIXTRANSFORM\_REMOVE\_er \\ 
			4953  & REMOVE\_es & \$SUFFIXTRANSFORM\_REMOVE\_es \\ 
			4954  & REMOVE\_est & \$SUFFIXTRANSFORM\_REMOVE\_est \\ 
			4955  & REMOVE\_ful & \$SUFFIXTRANSFORM\_REMOVE\_ful \\ 
			4956  & REMOVE\_ing & \$SUFFIXTRANSFORM\_REMOVE\_ing \\ 
			4957  & REMOVE\_ive & \$SUFFIXTRANSFORM\_REMOVE\_ive \\ 
			4958  & REMOVE\_less & \$SUFFIXTRANSFORM\_REMOVE\_less \\ 
			4959  & REMOVE\_ly & \$SUFFIXTRANSFORM\_REMOVE\_ly \\ 
			4960  & REMOVE\_n & \$SUFFIXTRANSFORM\_REMOVE\_n \\ 
			4961  & REMOVE\_ness & \$SUFFIXTRANSFORM\_REMOVE\_ness \\ 
			4962  & REMOVE\_y & \$SUFFIXTRANSFORM\_REMOVE\_y \\ 
						\hline
		\end{tabular}
	\end{center}
	\label{tab:suffix_trans_remove}
\end{table*}
\begin{table*}
	\caption{List of suffix append transformations.}
 	\small
	\begin{center}
		\begin{tabular}{l  ll }
			\hline
			id  & Transformation suffix & Tag \\ 
			\hline
			4906  & APPEND\_able & \$SUFFIXTRANSFORM\_APPEND\_able \\ 
			4907  & APPEND\_age & \$SUFFIXTRANSFORM\_APPEND\_age \\ 
			4908  & APPEND\_al & \$SUFFIXTRANSFORM\_APPEND\_al \\ 
			4909  & APPEND\_ation & \$SUFFIXTRANSFORM\_APPEND\_ation \\ 
			4910  & APPEND\_d & \$SUFFIXTRANSFORM\_APPEND\_d \\ 
			4911  & APPEND\_ed & \$SUFFIXTRANSFORM\_APPEND\_ed \\ 
			4912  & APPEND\_er & \$SUFFIXTRANSFORM\_APPEND\_er \\ 
			4913  & APPEND\_es & \$SUFFIXTRANSFORM\_APPEND\_es \\ 
			4914  & APPEND\_est & \$SUFFIXTRANSFORM\_APPEND\_est \\ 
			4915  & APPEND\_ful & \$SUFFIXTRANSFORM\_APPEND\_ful \\ 
			4916  & APPEND\_ing & \$SUFFIXTRANSFORM\_APPEND\_ing \\ 
			4917  & APPEND\_ist & \$SUFFIXTRANSFORM\_APPEND\_ist \\ 
			4918  & APPEND\_ive & \$SUFFIXTRANSFORM\_APPEND\_ive \\ 
			4919  & APPEND\_ly & \$SUFFIXTRANSFORM\_APPEND\_ly \\ 
			4920  & APPEND\_n & \$SUFFIXTRANSFORM\_APPEND\_n \\ 
			4921  & APPEND\_ness & \$SUFFIXTRANSFORM\_APPEND\_ness \\ 
			4922  & APPEND\_ship & \$SUFFIXTRANSFORM\_APPEND\_ship \\ 
			4923  & APPEND\_wise & \$SUFFIXTRANSFORM\_APPEND\_wise \\ 
			4924  & APPEND\_y & \$SUFFIXTRANSFORM\_APPEND\_y \\ 
						\hline
		\end{tabular}
	\end{center}
	\label{tab:suffix_trans_append}
\end{table*}

\begin{table*}
	\caption{Error patterns used for generating the pretraining dataset.}
	\small 
	\begin{center}
		\begin{tabular}{|p{2.5cm}|p{9cm}|p{3cm}|}
						\hline
			Name of the error & Options & Action \\
						\hline
			Change verb type & ['inf', '1sg', '2sg', '3sg', 'pl', 'part', 'p', '1sgp', '2sgp', '3sgp', 'ppl', 'ppart'] & Change a type with a random type from the list \\
			Prepositions  &  [ '', 'of', 'with', 'at', 'from', 'into', 'during', 'including', 'until', 'against', 'among', 'throughout','despite', 'towards', 'upon', 'concerning', 'to', 'in', 'for', 'on', 'by', 'about', 'like',    'through', 'over', 'before', 'between', 'after', 'since', 'without', 'under', 'within', 'along', 'following', 'across', 'behind', 'beyond', 'plus', 'except', 'but', 'up', 'out', 'around', 'down', 'off', 'above', 'near'] & Change a type with a random preposition from the list \\
			Determiners  & ['the', 'a', 'an', 'that', 'this',''] & Change a type with a random DET from the list\\
			Common letter patterns  &  \{'mb': 'm', 'bt': 't', 'tch': 'ch', 'tm': 'm', 'stle': 'sle', 'wh': 'w', 'hono': 'ono', 'hou': 'ou', 'hones': 'ones', 'rh': 'r', 'kn': 'n', 'sw': 's', 'wr': 'r', 'who': 'ho', 'gn': 'n', 'gu': 'g', 'ui': 'i', 'sc': 's', 'al': 'a', 'pn': 'n', 'ps': 's', 'pb': 'b', 'dg': 'g', 'dn': 'n', 'mn': 'm', 'isl': 'il', 'ough':'uf', 'through':'thro', 'though':'tho', 'ea':'ae', 'ei':'ie','au':'ua', 'exh':'ex','tion':'sion','sion':'tion','sure':'shure','cture':'cshre','ere':'ear','ear':'ere'\} & Switch keys with values\\
			Vowels Combinations  & ['ea','ou','ei','ie','ai','uo','io','oi','au','ua','ow','wo'] & Swap Vowels order \\
			Similar Sound  &  \{'a':['u'], 'b':['p'], 'p':['b'], 'e':['i','a'], 'o':['u','w'], 'f':['v'], 'w':['o','u'], 'u':['a','o','w'], 'i':['e','a','y'],'v':['f'],'y':['i']\} & Switch keys with values \\
			POS  & NN, NNS, VB, JJ, JJR, JJS, RB & Change a type with a random type from the list \\
						\hline
		\end{tabular}
\end{center}
\label{tab:error_pattterns}
\end{table*}

\label{}









\end{document}